\documentclass[sigconf]{acmart}
 
\usepackage{booktabs} % For formal tables
\usepackage{bm}
\usepackage{color}
\usepackage{subfigure}
\usepackage{multirow}

\def\D{\mathcal{D}} %domain
\def\a{\bm{a}} %category attribute words vector
\def\b{\bm{b}} %bias vector
\def\w{\bm{w}} %words vector
\def\W{\bm{W}} %filter
\def\L{\mathcal{L}}%loss function
\def\s{\bm{s}} %cosine similarity vector
\def\v{\bm{v}} %sentence level category feature vector
\def\m{\bm{m}} %feature map
\def\o{\bm{o}} %concat vector
\def\y{\bm{\hat{y}}} %prediction
\def\q{\bm{q}} %query words vector
\def\y{\bm{y}} %true label vector
\def\T{t} %each word's times in vocabulary
\def\X{\bm{X}} %domain samples collection
\def\x{\bm{x}} %domain sample
\def\model{\textbf{CAN-CNN }} %model name
\def\setR{\mathbb{R}}%
\setcopyright{acmcopyright}

\copyrightyear{2020}
\acmYear{2020} 
\setcopyright{iw3c2w3}
\acmConference[WWW '20]{Proceedings of The Web Conference 2020}{April 20--24, 2020}{Taipei, Taiwan} 
\acmBooktitle{Proceedings of The Web Conference 2020 (WWW '20), April 20--24, 2020, Taipei, Taiwan}
\acmPrice{}
\acmDOI{10.1145/3366423.3380088}
\acmISBN{978-1-4503-7023-3/20/04}

\begin{document}

\title{Domain Adaptation with Category Attention Network for Deep Sentiment Analysis}

\author{
Dongbo Xi$^{1,2}$,
Fuzhen Zhuang$^{1,2,*}$,
Ganbin	Zhou$^{3}$,
Xiaohu	Cheng$^{3}$,
Fen	Lin$^{3}$ and
Qing He$^{1,2}$
}
\affiliation{\institution{$^1$Key Lab of Intelligent Information Processing of Chinese Academy of Sciences (CAS),\\
Institute of Computing Technology, CAS, Beijing 100190, China}}
\affiliation{\institution{$^2$University of Chinese Academy of Sciences, Beijing 100049, China}}
\affiliation{\institution{$^3$Search Product Center, WeChat Search Application Department, Tencent, China}}
\email{{xidongbo17s, zhuangfuzhen, heqing}@ict.ac.cn, {ganbinzhou, xiaohucheng, felicialin}@tencent.com}

\thanks{* Corresponding author: Fuzhen Zhuang.}

\renewcommand{\shortauthors}{Xi and Zhuang, et al.}

\begin{abstract}
Domain adaptation tasks such as cross-domain sentiment
classification aim to utilize existing labeled data in the source domain and unlabeled or few labeled data in the target domain to improve the performance in the target domain via reducing the shift between the data distributions.
Existing cross-domain sentiment classification methods need to distinguish \textit{pivots}, i.e., the domain-shared sentiment words, and \textit{non-pivots}, i.e., the domain-specific sentiment words, for excellent adaptation performance. 
In this paper, we first design a Category Attention Network (CAN), and then propose a model named \model to integrate CAN and a Convolutional Neural Network (CNN).
On the one hand, the model regards \textit{pivots} and \textit{non-pivots} as unified category attribute words and can automatically capture them to improve the domain adaptation performance;
on the other hand, the model makes an attempt at interpretability to learn the transferred category attribute words.
Specifically, the optimization objective of our model has three different components: 
1) the supervised classification loss; 
2) the distributions loss of category feature weights; 
3) the domain invariance loss. 
Finally, the proposed model is evaluated on three public sentiment analysis datasets and the results demonstrate that \model can outperform other various baseline methods.
\end{abstract}




\maketitle

\section{Introduction}\label{introduction}
% Supervised tasks usually need large scale labeled data for outstanding performance.
Domain adaptation model can take the advantage of existing large scale labeled data in the source domain and adapt to unlabeled data with similar distribution in the target domain.
However, in sentiment classification, the expression of users' emotions usually varies across domains \cite{li2018hierarchical}.
For example, the word \textit{delicious} is used to express positive sentiment in the \textit{Foods} domain, whereas \textit{heartwarming} is used in the \textit{Movies} domain. Due to the domain discrepancy, a sentiment classifier trained in a source domain may not work well when directly applied to a target domain.

To address the problem, traditional methods need to
distinguish the domain-shared pivots and domain-specific non-pivots.
For example,
\citeauthor{blitzer2006scl1} \cite{blitzer2006scl1,blitzer2007scl2} proposed a Structural Correspondence Learning (SCL) method which
utilizes multiple pivot prediction tasks to infer the correlation between pivots and non-pivots.
% Similarly, \citeauthor{pan2010sfa} \cite{pan2010sfa} proposed a Spectral Feature Alignment (SFA) method to align the non-pivots with the pivots to build a bridge between the source and target domains. 
However, these methods need to deal with pivots and non-pivots separately, and they cannot be unified.
Recently, deep neural models are explored to reduce the shift in data distributions for domain adaptation.
Some studies utilized Maximum Mean Discrepancy (MMD) measure as a regularization to reduce the distribution mismatch between the source and target domains \cite{ghifary2014domain,tzeng2014deep,long2016unsupervised}. 
% Besides, some adversarial training based approaches~\cite{ganin2016domain,bousmalis2016domain,kim2017adversarial,cao2018partial,alam2018domain,shen2018wasserstein} have also gained popularity.
However, these deep models lack interpretability to directly learn the transferred category attribute words (i.e., pivots and non-pivots).

Along this line, we propose a \textbf{C}ategory \textbf{A}ttention \textbf{N}etwork and \textbf{C}onvolutional \textbf{N}eural \textbf{N}etwork based model (\textbf{CAN-CNN}).
On the one hand, the proposed model regards pivots and non-pivots as unified category attribute words to learn, which no longer needs to design different networks.
On the other hand, the category attribute words in target domain can be automatically learned to interpret what to transfer.
Specifically, the CAN module is the core of CAN-CNN and it contains a Category Memory Module (CMM), a Dynamic Matching (DM) process and a Category Attention (CA) layer.
We firstly construct a CMM which contains some category attribute words.
Then, a DM process can dynamically match category attribute words from the CMM for each sample.
Moreover, the CA acts on each category attribute word and each sample to pay more attention to category features in the sample.
Finally, we apply CAN to the source domain where CMM is specially constructed, and the target domain where CMM is randomly initialized to interpret what can be transferred from the source domain to the target domain.
Also, for obtaining significant performance improvements and improving the interpretability, the optimization objective of our model has three different components as described in Section \ref{allloss}.
By optimizing the objective of our model,
the proposed CAN can focus on category features and ignore non-category features in a sample for better performance in target domain, 
and the domain-aware CMM and the CA in the CAN can give inspiration on the interpretability in domain adaptation.
Comprehensive experimental results on real-world datasets demonstrate that the proposed \model can outperform other various baselines, 
and the further analysis of the experimental results demonstrates that the domain-aware CMM in CAN can be seen as a kind of knowledge transfer between source and target domain which can interpret what can be transferred from the source domain to the target domain.

\section{Related Work}
Traditional methods need to distinguish the pivots and non-pivots.
For example, Structural Correspondence Learning (SCL) method \cite{blitzer2006scl1,blitzer2007scl2} 
induce a low-dimensional feature representation shared across domains based on the co-occurrence between pivots and non-pivots, and it needs to utilize multiple pivot prediction tasks to infer the correlation between pivots and non-pivots.
Similarly, the Spectral Feature Alignment (SFA) method \cite{pan2010sfa} aims to align the non-pivots from different domains into unified clusters,
with the help of pivots as a bridge.
% Besides, the Adversarial Memory Network (AMN) \cite{li2017end} can automatically
% capture the pivots using an attention mechanism, but it cannot automatically
% capture and exploit non-pivots, which may result in the degraded performance when source and target domains only have few overlapping pivot features.
Besides, the Hierarchical Attention Transfer Network
(HATN) \cite{li2018hierarchical} can automatically capturing pivots and non-pivots simultaneously, but it needs to utilize a P-net and a NP-net to deal with pivots and non-pivots separately, and they cannot be unified. 

Recently, deep learning methods form another line of
work to automatically produce superior feature representations for cross-domain sentiment classification.
\citeauthor{glorot2011domain} \shortcite{glorot2011domain} utilized Stacked Denoising Autoencoder (SDA) to extract a meaningful representation in an unsupervised fashion and then trained a classifier with this high-level feature representation.
Other leading works usually utilize Maximum Mean Discrepancy (MMD) measure as a regularization to reduce the distribution mismatch between the source and target domains \cite{ghifary2014domain,tzeng2014deep,long2015learning,long2016unsupervised}. 
% \citeauthor{long2015learning} \shortcite{long2015learning} applied the multiple kernel variant of MMD (MK-MMD) proposed by \citeauthor{gretton2012optimal} \shortcite{gretton2012optimal} to CNN to reduce the dataset bias and enhance the transfer ability in task-specific layers.
% Besides, \citeauthor{dong2018helping} \shortcite{dong2018helping} proposed to induce sentiment embedding via supervision on out-of-domain data, which is then fed into the model via a dedicated memory based component.
Other methods that have gained popularity recently are utilizing adversarial training or Generative Adversarial Nets (GAN) \cite{goodfellow2014generative}. These core approaches include Domain Adaptation with Adversarial Training (DAAT) \cite{alam2018domain},
Domain-Adversarial Neural Network (DANN) \cite{ganin2016domain}, Domain Separation Networks (DSN) \cite{bousmalis2016domain}, Selective Adversarial Networks (SAN) \cite{cao2018partial} and some other approaches \cite{liu2016coupled,kim2017adversarial,shen2018wasserstein}. 
However, these studies usually lack interpretability to directly learn the transferred category attribute words,
and their poor interpretability in domain adaptation causes trouble for user understanding and makes the model performance untrustworthy when applied to real applications.

\section{Methodology}
\begin{figure*}[!t]
\begin{center}
\includegraphics[width=0.78\linewidth]{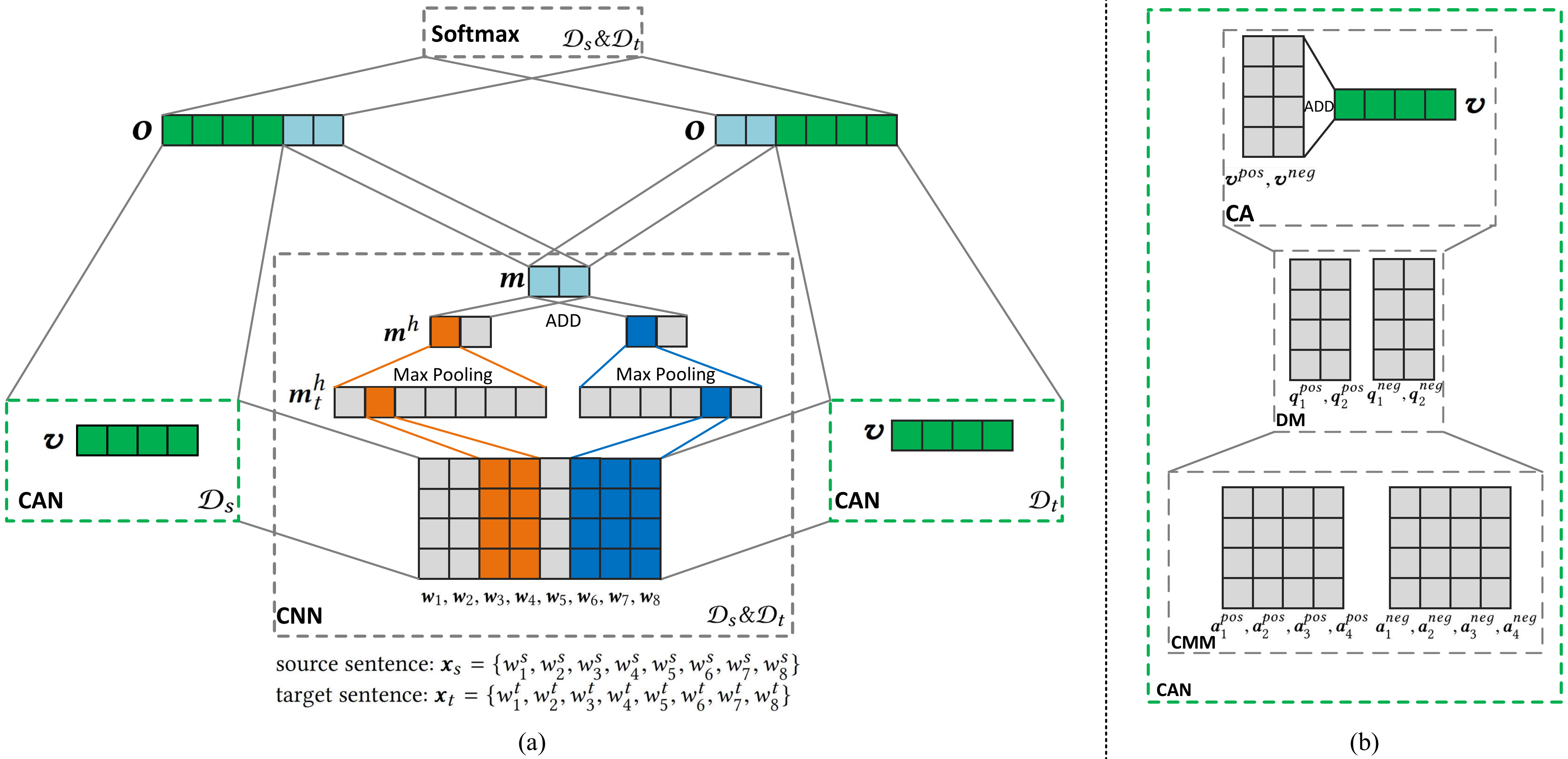}
\caption{(a) The CAN and CNN based domain adaptation model \model.  (b) The CAN module details for binary classification sentiment analysis task.}
\label{fig:model}
\end{center}
\end{figure*}

In this section, we first formulate the problem, then present the details of the proposed model \model as shown in Figure \ref{fig:model}(a).

\subsection{Problem Formulation}
Given a source domain $\D_s$ and a target domain $\D_t$. Suppose the source and target domain both have a classification task whose number of categories is $C$.
Let $\X_s = \{(\x_1^s,y_1^s),\cdots,(\x_{N_s}^s,y_{N_s}^s)\}$ is a set of $N_s$ labeled samples in the source domain, where $\x_i^s\in\setR^L$ is the $i$-th sample with length $L$ in the source domain, $y_i^s\in\{1,2,\cdots,C\}$ is the label of $i$-th  sample $\x_i^s$. 
Similarly, $\X_t = \{\x_1^t,\cdots,\x_{N_t}^t\}$ is a set of $N_t$ unlabeled samples in the target domain, where $\x_j^t\in\setR^L$ is the $j$-th target domain sample with length $L$.
The task is to utilize labeled training data $\X_s^{train}$ in the source domain and unlabeled training data $\X_t^{train}$ in the target domain to train a model, which has a good classification performance on the target data $\X_t^{test}$.
Here, for simplicity, we will discuss only binary classification ($C=2$) task here. But our model can be adapted to multi-class tasks.

\subsection{Category Attention Network (CAN)}

Category features (e.g. words with category attributes) play an important indicative role in classification tasks. 
Taking sentiment analysis as an example, some positive sentiment words (e.g., \textit{funny}, \textit{pleasing}) occur frequently in positive text, while some negative sentiment words (e.g., \textit{boring}, \textit{disgusting}) are more common in negative text. 
Therefore, we design the \textbf{CAN}, as shown in Figure \ref{fig:model}(b), to focus on category features and ignore no-category features. 
As mentioned in Section \ref{introduction}, the CAN module is the core of CAN-CNN and it contains a Category Memory Module  (CMM), a Dynamic Matching (DM) process and a Category Attention (CA) layer.

\subsubsection{Category Memory Module (CMM)}\label{sec:cmm}
The CMM in source domain contains some category attribute words drawn from the labeled data. 
Different categories should have different category attribute words and each category's attribute words should occur frequently in this category but rarely seen in other categories.
Therefore, for the labeled source domain data, we extract the positive category attribute words as follows:
\begin{equation}
    a_1^{pos} = \mathop{\arg\max}_{i}\left(\log\T^{pos}_i-\log\T^{neg}_i\right)\label{equ:cmm},
\end{equation}
where $\T^{pos}_i$ and $\T^{neg}_i$ are the numbers of $i$-th word in the vocabulary appears in the positive samples and negative samples, respectively.
According to Equation (\ref{equ:cmm}), we can get the positive top-$1$ attribute word $a_1^{pos}$. Besides, we can also get the negative top-$1$ attribute word $a_1^{neg}$ by replacing $\arg\max$ with $\arg\min$ in Equation (\ref{equ:cmm}).
The CMM can be constructed via selecting top-$M$ attribute words for each category. Therefore, the CMM should contain $M\times 2$ category attribute words which are $M$ positive words $\{a_1^{pos},\cdots,a_M^{pos}\}$ and $M$ negative words $\{a_1^{neg},\cdots,a_M^{neg}\}$.

Because our proposed CMM needs to utilize labeled data to extracted category attribute words, it can't be directly utilized on the unlabeled target domain data. 
Therefore, we need to apply a domain-aware CMM.
The domain-aware CMM contains pivots and non-pivots within a unified framework.
In our model, the CMM in the target domain should be randomly initialized word vectors and can be learned automatically during the model training by minimizing the Distributions Loss of Category Feature Weights in Section \ref{cmmloss}.
Finally, we hope the learned CMM in the target domain can carry category information which can be seen as a kind of knowledge transfer between the source and target domains and can improve the interpretability.

\subsubsection{Dynamic Matching (DM)}
In the word embedding space, the DM process utilizes cosine similarity to dynamically match the top-$K$ most similar category attribute words from the CMM for each sample during each gradient-descent step as follows:
\begin{eqnarray}
    s_m^c&=&\max_{l\in \{1,\cdots,L\}}cos(\w_l, \a_m^c)\label{equ:query1},\\
    q_1^c&=&\mathop{\arg\max}_{m\in \{1,\cdots,M\}}s_m^c\label{equ:query2},
\end{eqnarray}
where $\w_l\in\setR^d$ is the embedding vector of $l$-th word in sentence $\x$, $\a_m^c\in\setR^d$ is the embedding vector of $m$-th category attribute word in category $c$ ($c\in\{pos,neg\}$), and $d$ is the dimension of the embedding vector.
According to Equation (\ref{equ:query2}), we can get the top-$1$ matched category attribute word $q_1^c$ for category $c$. The DM process dynamically selects top-$K$ matched category attribute words from $M$ category attribute words for each category ($K\ll M$). Therefore, we finally get $K\times 2$ matched category attribute words, i.e., $K$ positive words $\{q_1^{pos},\cdots,q_K^{pos}\}$ and $K$ negative words $\{q_1^{neg},\cdots,q_K^{neg}\}$. These matched category attribute words can indicate the sentiment polarity of the sentence $\x$.

\subsubsection{Category Attention (CA)}\label{sub:CA}
Attention mechanism has been proved to be effective in machine translation \cite{bahdanau2014neural}, where the attention makes the model focus on useful words for the current task. Hence, we utilize CA to act on each matched category attribute words $q_k^c$ and the sentence $\x$ to pay more attention to the category features in the sentence. 
We compute the weight vector $\s_{k}^c\in\setR^L$ between category attribute word $\q_k^c$ and sentence $\x$ as:
\begin{eqnarray}
    s_{k,l}^c=\frac{\exp\left( \q_k^{c \top}\tanh(\W_a\w_l+\b) \right)}{\sum_{l=1}^L \exp\left(\q_k^{c \top}\tanh(\W_a\w_l+\b)\right)}\label{equ:weight},
\end{eqnarray}
where $\W_a\in\setR^{d\times d}$ and $\b\in\setR^{d}$ are the parameters of weight and bias of the category attention, respectively.
The degree of correspondence between $\q_k^c$ and $\w_l$ can be assessed by the Equation (\ref{equ:weight}). The resulting vector of scores $\s_k^c$ yields category feature weights for each word in the sentence.
The output sentence-level category feature vector $\v^c$ for  category $c$ is that first a sum over the sentence inputs $\x$ weighted by the scores vector $\s_k^c$ and then a sum over $K$ weighted vectors $\v_{k}^{c}$:
$
    \v^c=\sum_{k=1}^K\sum_{l=1}^L s_{k,l}^c \w_l.
$

Finally, We sum the sentence-level category features $\{\v^{pos}, \v^{neg}\}$ to get the overall feature representation:
$
  \v=\v^{pos}+\v^{neg}.
$

\subsection{CAN and CNN based Domain Adaptation} \label{allloss}
Convolutional Neural Network (CNN) has been proved to be efficient on text classification~\cite{kim2014convolutional}, and recently has been applied to transfer text classification~\cite{dong2018helping,alam2018domain}.
We apply the same CNN architecture as the TextCNN \cite{kim2014convolutional} (as shown in Figure \ref{fig:model}(a)) which is an embedding layer, followed by a convolution layer with $T$ filters of different width $h$, and then a summation (or concatenation) of global max pooled features to generate the final features $\m$.

The CAN learned  features $\v\in\setR^d$ and the CNN learned features $\m\in\setR^T$ are concatenated, and then we apply a fully-connected layer with softmax activation to output the final prediction:
\begin{eqnarray}
  \o&=&[\v;\m],\label{equ:last2}\\
  \hat{\y}&=&\mathrm{softmax}\left(\W_s^{\top}\o\right),
\end{eqnarray}
where $[\cdot ; \cdot]$ denotes the concatenation of two column vectors, $\W_s^{\top}\in\setR^{C\times(d+T)}$ 
are the weights of fully-connected layer, and $\hat{\y}\in\setR^C$ is a distribution which indicates different category probability. 

The source domain and target domain data share all network structure and parameters except the domain-aware CMM.
The loss function $\L(\theta)$ of our model is split into three components each of which plays a different role:
\begin{align}
  \L(\theta)=\L_C(\theta)+\alpha\L_D(\theta)
  +\beta \L_I(\theta),\label{equ:loss}
\end{align}
where $\theta$ is the parameters set, the tunable hyper-parameters $\alpha$ and $\beta$ control the different strength of the components.
Training is finished through stochastic gradient descent over shuffled mini-batches with the Adam \cite{kingma2014adam} update rule.

\subsubsection{Supervised Classification Loss}
The labeled data $\X_s$ in the source domain $\D_s$ are available for our classification task. We need to minimize the \textit{cross entropy} of predicted distribution and the actual distribution: 
\begin{equation}
\L_C(\theta)=-\frac{1}{N_s}\sum_{\D_s}\sum^C_{c=1}y_{i,c}^s\log(\hat{y}_{i,c}^s\vert \x_i^s,\theta),
\end{equation}
where $N_s$ is the number of source domain samples, $C$ is the number of categories, $\y_i^s\in\setR^C$ and $\hat{\y}_{i}^s\in\setR^C$ are the one-hot label and the predicted distribution of source domain sample $\x_i^s$, respectively, and $\theta$ is the parameters set.

\subsubsection{Distributions Loss of Category Feature Weights} \label{cmmloss}
The different category attribute words should pay different attention to different category features. For example, positive category attribute words should pay more attention to positive words 
(e.g., \textit{happy}, \textit{pleased}, $\cdots$) in a sentence while pay less attention to the negative words (e.g., \textit{boring}, \textit{unfunny}, $\cdots$). 
Therefore, distribution $\s_k^c$ of category feature weights should be discrepant for different categories. We need to minimize negative Kullback-Leibler (KL) divergence \cite{kullback1987letter} between positive and negative categories. Here, the constraint should be applied to both source and target domains:
\begin{small}
\begin{align}
\L_D(\theta)=&-\frac{1}{N_s}\sum_{\D_s}\left(\mathrm{KL}\left(\s^{pos},\s^{neg}\right)+\mathrm{KL}\left(\s^{neg},\s^{pos}\right)\right)\nonumber\\
&-\frac{1}{N_t}\sum_{\D_t}\left(\mathrm{KL}\left(\s^{pos},\s^{neg}\right)+\mathrm{KL}\left(\s^{neg},\s^{pos}\right)\right)\label{equ:2kl},
\end{align}
\end{small}
where
$
\s^c=\sum_{k=1}^K\s^c_k
$
and
$
\s_k^{c}=\{s_{k,1}^{c},\cdots,s_{k,L}^{c}\}
$
are the distribution of category feature weights for category $c$ and $k$-th matched category attribute word, respectively.
By minimizing the Equation (\ref{equ:2kl}), the attention distributions of different category attribute words are pulled apart. This constraint can force the target domain to yield domain-aware CMM which can be seen as a kind of knowledge transfer to interpret what can be transferred from the source domain to the target domain.
Utilizing the domain-aware CMM, the model can increase the  performance in the target domain.

\subsubsection{Domain Invariance Loss}
Due to the distribution discrepancy between source and target domains, 
directly training a model utilizing only the source data usually leads to overfitting to the source data 
distribution~\cite{tzeng2014deep}. It also reduces the performance when it predicts the label in the target domain. 
Our model expects to reduce the distribution mismatch between source and target domains in the latent space to learn domain invariant features and provide a good domain adaptation performance.%, as shown in Figure \ref{fig:mmd}.
To minimize this discrepancy, we adopt the Maximum Mean Discrepancy (MMD)~\cite{borgwardt2006integrating} as 
the metric. MMD is a widely used regularization method in image domain 
adaptation~\cite{tzeng2014deep,ghifary2014domain,long2015learning}. 
The MMD is performed on the source and target domains:
\begin{equation}
  \L_I(\theta)=\Vert\frac{1}{N_s}\sum_{\D_s}\phi(\x_s)-\frac{1}{N_t}\sum_{\D_t}\phi(\x_t)\Vert,
\end{equation}%
this distance is computed with respect to a particular representation, $\phi(\cdot)$, and this representation operates on source data point $\x_s$ and target data point $\x_t$. In our case, it is an empirical approximation to this distance. We apply the MMD to the penultimate layer of our model as performed in~\cite{tzeng2014deep}. 
By minimizing the classification loss $\L_C(\theta)$ and the domain invariance loss $\L_I(\theta)$ in Equation (\ref{equ:loss}) simultaneously, the model can remain invariant to domain shift and yield a strong classification representation.

It is worth noting that our approach is a general framework and the optimization objective acts on the 
CAN and the features extractor CNN. So the CNN can be replaced by any other efficient feature extractors 
(e.g., LSTM \cite{hochreiter1997long}, Transformer \cite{Vaswani2017Attention}).

\begin{table}[!t]
  \centering
  \caption{The extracted top-$5$ category attribute words for positive class and negative class by CMM on three different domain datasets CR, AFF and MR.}
  %\resizebox{1\linewidth}{0.16\linewidth}{
  %\setlength{\tabcolsep}{0.5mm}{
  \scalebox{0.85}{
    \begin{tabular}{cccccc}
    \toprule
     \multicolumn{2}{c}{\textbf{CR}} & \multicolumn{2}{c}{\textbf{AFF}} &
     \multicolumn{2}{c}{\textbf{MR}}\\
    pos. & neg. & pos. & neg. & pos. & neg. \\
    \midrule
    love & useless & Delicious & worst & engrossing & badly \\
    happy & flaw & beat & terrible & realistic & unfunny \\
    pleased & press & Perfect & worse & riveting & tiresome \\
    amazing & recognize & Excellent & horrible & gem & disguise \\
    great & downside & fast & disgusting & heartwarming & boring \\
    \bottomrule
    \end{tabular}
    }
  \label{tab:cmm}%
\end{table}%

\begin{table*}[!t]
  \centering
  \caption{Accuracy based on 10-fold cross validation on different target domain datasets transferred from different source domain datasets, comparing our \model model against various baselines.}
  \scalebox{0.95}{
  %\resizebox{1\linewidth}{0.186\linewidth}{
    \begin{tabular}{clcccccc}
    \toprule
        & \textbf{Model} & 
         \textbf{MR}$\rightarrow$\textbf{CR} &
         \textbf{AFF}$\rightarrow$\textbf{CR} &
         \textbf{CR}$\rightarrow$\textbf{AFF} & \textbf{MR}$\rightarrow$\textbf{AFF} & \textbf{CR}$\rightarrow$\textbf{MR} & \textbf{AFF}$\rightarrow$\textbf{MR} \\
    \midrule
    \multirow{5}[2]{*}{Direct Transfer} & fastText-rand  & 0.6250 & 0.6700 & 0.6738  & 0.6878  & 0.5727  & 0.5824  \\
        & fastText-non-static  & 0.6674 & 0.7462 & 0.7232  & 0.7476  & 0.6535  & 0.6878  \\
        & CNN-char  & 0.5588 & 0.6658 & 0.7139  & 0.6608  & 0.5603  & 0.5923  \\
        & CNN-rand  & 0.6052 & 0.6985 & 0.7117  & 0.6731  & 0.5884  & 0.5996  \\
        & CNN-non-static  & 0.6884 & 0.7578 & 0.7517  & 0.7624  & 0.6679  & 0.6918  \\
    \midrule
    \multicolumn{1}{c}{\multirow{6}[3]{*}{Domain Adaptation}} 
    & SDA   & 0.6058 & 0.6624 & 0.6742 & 0.6920 &0.6238 & 0.6324\\
    & mSDA   & 0.5954 & 0.6426 & 0.6803 & 0.7051 &0.6206 & 0.6387\\
    & SDA-fine-tune   & 0.6224 & 0.6939 & 0.6893 & 0.7143 &0.6305 & 0.6422\\
    & DAAT  & 0.6981 & 0.7302 & 0.7211  & 0.7433  & 0.6235  & 0.6521  \\
\cmidrule{2-8}        
        & \textbf{CAN-CNN}-share  & 0.7149 & 0.7493  & 0.7654  & 0.7808  & 0.6548  & 0.6960  \\
        & $\model$ & \textbf{0.7302} & \textbf{0.7647} & \textbf{0.7882}  & \textbf{0.7929}  & \textbf{0.6796} & \textbf{0.7098} \\
    \bottomrule
    \end{tabular}%
    }
  \label{tab:result}%
\end{table*}%

\section{Experiments}
In this section, we perform experiments to evaluate the proposed model against various baselines on three real-world datasets. 
% We first introduce the datasets, implementation details and baseline methods of our experiments. Finally, we present our experimental results and analysis.

\subsection{Datasets}
We utilize three public sentiment analysis datasets in the experimental study. 

1) \textbf{CR}\footnote{http://www.cs.uic.edu/$\sim$liub/FBS/sentiment-analysis.html}~\cite{hu2004mining} is a customer review sentiment analysis dataset of various products (e.g., cameras, DVD player, MP3 etc.) from amazon.

2) \textbf{AFF}\footnote{http://snap.stanford.edu/data/web-FineFoods.html}~\cite{mcauley2013amateurs} is a sentiment dataset consisting of reviews of fine foods from amazon. We chose a random subset of it as done in \cite{dong2018helping}.

3) \textbf{MR}\footnote{https://www.cs.cornell.edu/people/pabo/movie-review-data/}~\cite{pang2005seeing} is a movie review sentiment analysis dataset.
%\end{itemize}

As there is no standard train/test split for the three datasets, we adopt 10-fold cross validation.
The extracted top-$5$ category attribute words for positive class and negative class by CMM as Equation (\ref{equ:cmm}) on three source domain datasets CR, AFF and MR are shown in Table \ref{tab:cmm}.
And the results further show that different domains should have different category attribute words even on the same category, 
so we can not directly apply the source domain CMM to the target domain, we need to apply a domain-aware CMM.

\begin{table}[!t]
  \centering
  \caption{Target domain accuracy of ablation study.}
  \scalebox{0.85}{
  %\resizebox{1\linewidth}{0.186\linewidth}{
    \begin{tabular}{lccc}
    \toprule
    \textbf{Model} & 
         \textbf{MR}$\rightarrow$\textbf{CR} &
         \textbf{CR}$\rightarrow$\textbf{AFF} & \textbf{AFF}$\rightarrow$\textbf{MR} \\
 \midrule
 $\model(\alpha=0, \beta=0)$ 
  & 0.7164  & 0.7661    & 0.7008  \\
  $\model(\alpha=0)$  & 0.7281 & 0.7700   & 0.7044  \\
 $\model(\beta=0)$  & 0.7148  & 0.7867   & 0.6989  \\
$\model$ & \textbf{0.7302}  & \textbf{0.7882}  & \textbf{0.7098} \\
    \bottomrule
    \end{tabular}%
    }
  \label{tab:ablation}%
\end{table}%

\subsection{Implementation Details}
We can construct 6 domain adaptation tasks based on the three different datasets: MR$\rightarrow$CR, AFF$\rightarrow$CR, CR$\rightarrow$AFF, MR$\rightarrow$AFF, CR$\rightarrow$MR, AFF$\rightarrow$MR,

For fair comparison, for all datasets we use: filter width ($h$) of 3, 4, 5 with 100 ($T$) feature maps each and embedding dimension of 300 ($d$) which are same as the setting in the TextCNN. Besides, we use: mini-batch size of 128 and learning rate of 0.001. These hyper-parameters are the same for all datasets.

In the CAN module, for all datasets we use: $M=50$ category attribute words for each category in CMM, $K=5$ dynamically matched category attribute words for each category in DM, $\alpha=0.05$ and $\beta=0.01$ as the strength of the different components in the loss function. These values are chosen via a grid search on the CR validation set which splits $10\%$ from training set.

In all experiments, we adopt \textbf{Accuracy} as our evaluation metric based on 10-fold cross validation.
% We do not perform any dataset-specific tuning except early stopping on validation sets. 

\subsection{Baselines}
We compare the proposed method \model with two kinds of baselines. The one is direct transfer baselines following the existing work~\cite{alam2018domain}: we train a model (e.g., fastText, CNN) on the source domain and transfer it directly to the target domain. The other one is deep domain adaptation baselines which do not need to capture pivots and non-pivots separately.

1) \textbf{direct transfer baselines}: 
We adopt fastText~\cite{joulin2017bag} with randomly initialized word vectors \textbf{fastText-rand} and pre-trained fine-tuned word vectors \textbf{fastText-non-static} as our baselines. We also adopt TextCNN~\cite{kim2014convolutional} with randomly initialized word vectors \textbf{CNN-rand} and pre-trained fine-tuned word vectors \textbf{CNN-non-static} as our baselines. Besides, character-level~\cite{zhang2015character} TextCNN \textbf{CNN-char} is also adopted in our experiment.
We initialize the word embedding using a open-source pre-trained data \cite{mikolov2013distributed}\footnote{https://code.google.com/p/word2vec/}.

2) \textbf{domain adaptation baselines}:
For domain adaptation,
we adopt \textbf{SDA} \cite{glorot2011domain} and \textbf{mSDA} \cite{chen2012marginalized} with a SVM classifier as the baselines.
Also, we initialize the network architecture and parameters with the learned SDA, and fine-tune (\textbf{SDA-fine-tune}) with a softmax classifier on the source domain data.
To be fair to compare, we also adopt \textbf{DAAT}~\cite{alam2018domain} which is a domain adaptation method with adversarial training without the semi-supervised component. 

\subsection{Performance Comparison}
The experimental results evaluated by Accuracy based on 10-fold cross validation on different target domain datasets are presented in Table \ref{tab:result}. 
CNN-char performs the worst due to the lack of word-level information.
CNN-rand improves the results comparing with CNN-char by utilizing word-level information, and fastText-rand obtains similar  improvement.
Utilizing pre-trained word vectors can further improve the performance on target domain datasets (fastText-non-static vs fastText-rand, CNN-non-static vs CNN-rand). It makes sense that the pre-trained word vectors which are trained on large-scale dataset utilize out-of-domain knowledge to reduce the overfitting risk and can give a better transfer performance.

For domain adaptation baselines, SDA and mSDA obtain similar performance, and SDA-fine-tune outperforms SDA via fine-tuning the model parameters.
Besides, CNN-non-static is the best method among the baselines including the domain adaptation model DAAT. DAAT utilizes adversarial training which is difficult to converge and does not perform good enough on the sentiment analysis task.

And we also let the CMM in target domain be the same as the CMM in source domain (\textbf{CAN-CNN}-share) to demonstrate the effectiveness of the domain-aware CMM. 
We can see that \textbf{CAN-CNN}-share performs worse than \model due to utilizing shared CMM not the domain-aware CMM.
This further illustrates that we can not directly apply the source domain CMM to the target domain.
On the contrary, \model obtains the best performance in multiple pairs experiments via utilizing domain-aware CMM.
These improvements indicate that the proposed \model model can better handle the domain adaptation task.
It is worth noting that all the experiments are unsupervised in the target domain, and if we utilize a small amount of labeled samples in the target domain, the performance of our proposed model will be further improved.

Furthermore, for understanding the contribution of different components in loss function in Equation (\ref{equ:loss}), we mask different components to demonstrate the effectiveness: 1) Only supervised loss \model$(\alpha,\beta=0)$; 2) With domain invariance loss \model$(\alpha=0)$; 3) With distributions loss of category feature weights \model$(\beta=0)$. 
As shown in Table \ref{tab:ablation}, we can see that the \model
obtains the best performance in multiple pairs experiments, which indicates each component matters for the performance.

\begin{table}[!t]
  \centering
  \caption{The words in the vocabulary most similar to category attribute words in the target domain CMM before (random initialization) and after (fine-tuning) training, respectively.}
  %\resizebox{1\linewidth}{0.2\linewidth}{
  \scalebox{0.81}{
  \setlength{\tabcolsep}{0.2mm}{
    \begin{tabular}{ccccccc}
    \toprule
    &\multicolumn{2}{c}{\textbf{MR}$\rightarrow$\textbf{CR}} &
        \multicolumn{2}{c}{\textbf{CR}$\rightarrow$\textbf{AFF}} &
        \multicolumn{2}{c}{\textbf{AFF}$\rightarrow$\textbf{MR}}\\
        & Before & After & Before & After & Before & After\\
        \midrule
            \multirow{5}[2]{*}{pos.}  & indicated & great & buck & nice & dolgin & heartwarming \\
         & sizing & wonderful & semi & delicious & programs & dramatize\\
         & took & amazing & United & good & adults & satiric\\
         & Lithium & vividly& diagnoised & simple  & cloying & screamingly \\
        & multiple & repeatly & shift & perfect & reawaken & vividly\\
    \midrule
    \multirow{5}[2]{*}{neg.}  & had & unfortunately  & addresses & unfortunately & adaptation & but \\
         & rebooting & useless & unclutterd & worse & assert & badly\\
         & 4mp & supposed & Shoprite & awful & hope & unfortunately\\
         & glow & flaw & toasted & a & plying & boring \\
        & stamp & awful & weight & terrible& niccol & loveless\\
    \bottomrule
    \end{tabular}
    }}
  \label{tab:interpretable}
\end{table}%
\subsection{Interpretability}\label{interpret}
As mentioned in Section \ref{sec:cmm}, our proposed CMM needs to utilize labeled data to extract category attribute words, so it can't be directly utilized on the unlabeled target domain data. 
Besides, we can observe (see Table \ref{tab:cmm}) that different domains have different category attribute words even on the same category (e.g., ``delicious", ``fast" for positive class in AFF, while ``heartwarming", ``vividly" for positive class in MR), so we can not directly apply the source domain CMM to the target domain, and the experiment results of model \textbf{CAN-CNN}-share in Table \ref{tab:result} also show the poor performance comparing with \textbf{CAN-CNN}. 
Therefore, in our model, the CMM in the target domain needs to be randomly initialized and can be learned automatically during the model training.
Table \ref{tab:interpretable} shows the words in the vocabulary most similar to category attribute words in the target domain CMM before (random initialization) and after (fine-tuning) training, respectively. 
Before training, the words most similar to category attribute words are some random words due to randomly initialization. 
After training, the most similar words obviously carry category information although with few noise words. 
It's because we minimize the Equation (\ref{equ:2kl}), the attention distributions of positive and negative category attribute words are pulled apart. This constraint can force the target domain to yield domain-aware CMM which can be seen as a kind of transfer of knowledge between source and target domain,
and the domain-aware CMM contains pivots and non-pivots within a unified framework.
Therefore, the proposed \model has the capability to interpret what can be transferred from the source domain to the target domain.

\begin{table}[!t]
  \centering
  \caption{The visualization of category attention weights on target domain MR.}
  \scalebox{0.72}{
    \begin{tabular}{clllll}
    \toprule
        & \multicolumn{5}{c}{Sentences} \\
    \midrule
    \multirow{2}[2]{*}{pos.} & \multicolumn{1}{l}{(1)}&
    \multicolumn{4}{l}{
    \colorbox[rgb]{ .988,  .988,  1}{an}
    \colorbox[rgb]{ .988,  .988,  1}{extremely}
    \colorbox[rgb]{ .388,  .745,  .482}{funny},
    \colorbox[rgb]{ .988,  .988,  1}{ultimately}
    \colorbox[rgb]{ .922,  .961,  .941}{heartbreaking}
    \colorbox[rgb]{ .949,  .973,  .969}{look}
    \colorbox[rgb]{ .998,  .998,  1}{at}
    \colorbox[rgb]{ .998,  .998,  1}{life}
    ...}
    \\ & \multicolumn{1}{l}{(2)}&
    \multicolumn{4}{l}{
    \colorbox[rgb]{ .529,  .804,  .604}{beautifully}
    \colorbox[rgb]{ .988,  .988,  1}{observed},
    \colorbox[rgb]{ .388,  .745,  .482}{miraculously}
    \colorbox[rgb]{ .98,  .988,  .992}{unsentimental}
    \colorbox[rgb]{ .988,  .988,  1}{comedy}
    \colorbox[rgb]{ .988,  .988,  1}{drama}.} \\
    \midrule
    \multirow{2}[2]{*}{neg.} & \multicolumn{1}{l}{(1)}&
    \multicolumn{4}{l}{
    \colorbox[rgb]{ .953,  .976,  .973}{godawful}
    \colorbox[rgb]{ .388,  .745,  .482}{boring}
    \colorbox[rgb]{ .89,  .949,  .914}{slug}
    \colorbox[rgb]{ .976,  .984,  .992}{of}
    \colorbox[rgb]{ .98,  .984,  .992}{a}
    \colorbox[rgb]{ .988,  .988,  1}{movie}.} \\
    & \multicolumn{1}{l}{(2)}&
    \multicolumn{4}{l}{        
    \colorbox[rgb]{ .796,  .91,  .831}{certainly}
    \colorbox[rgb]{ .678,  .863,  .733}{not}
    \colorbox[rgb]{ .98,  .984,  .992}{a}
    \colorbox[rgb]{ .62,  .839,  .682}{good}
    \colorbox[rgb]{ .988,  .988,  1}{movie},
    \colorbox[rgb]{ .886,  .949,  .914}{but}
    \colorbox[rgb]{ .824,  .922,  .859}{it}
    \colorbox[rgb]{ .941,  .969,  .961}{wasn't}
    \colorbox[rgb]{ .388,  .745,  .482}{horrible}
    \colorbox[rgb]{ .776,  .906,  .82}{either}.} \\
    \bottomrule
    \end{tabular}
    }
  \label{tab:case}%
\end{table}%
\subsection{Case Study}\label{case_interpret}
For two positive class samples and two negative class samples from target domain MR, we average the weights calculated by the top-$K$ matched category attribute words (see Equation (\ref{equ:weight})) as $\s^c=\frac{1}{K}\sum_k\s_k^c$ and visualize the corresponding category attention weights as shown in Table \ref{tab:case}.
The depth of color for each word illustrates the distributions of category attention weights. The darker the color is, the higher attention weight it has.

First, for the positive sample (1), we can see that the third word ``funny'' has the highest attention weight, and the word ``heartbreaking'' has a second highest attention weight. They both tell positive sentiment features which are corresponding to the actual label. 
In the positive sample (2), the similar case can also be observed. 
On the other hand, the negative sentiment features word ``unsentimental'' has a very small weight, which shows that the matched positive category attribute words can mask the fake sentiment information and pay more attention to the actual sentiment information. 

Second, for the negative sample (1), most of the weight is held by the second word ``boring'' which brings negative sentiment features, and the word ``slug'' also shares some weight due to its negative sentiment features. Besides, since there is a space missing between ``god'' and ``awful'' in the raw data, the matched category attribute words do not notice the negative sentiment information very well.
For the negative sample (2), the fourth word ``good'' brings positive sentiment features, while the tri-gram ``not a good'' tells negative sentiment features, the matched negative category attribute words can notice this, and pay more attention to the tri-gram. 
Besides, although the bi-gram ``wasn't horrible'' tells non-negative sentiment features, the whole sentence sentiment is negative. As we can see, our model is not masked by the bi-gram, but rather pays more attention to the negative sentiment word ``horrible''.

Such above results demonstrate that our \model model can effectively find the important parts within a sample according to the matched category attribute words for better performance. 

\section{Conclusion}
In this paper, we proposed a \model model which not only improves the domain adaptation performance, 
but also makes an attempt at interpretability via designing a Category Attention Network integrated with a Convolutional Neural Network.
Specifically, the proposed model regards pivots and non-pivots as unified category attribute words to learn, which no longer needs to design different networks, and the category attribute words in target domain can be learned automatically to interpret what to transfer.
The experimental results on three public sentiment analysis datasets from three domains 
demonstrate the performance improvements of our proposed model compared with various baseline methods. 

\begin{acks}
The research work supported by the National Key Research and Development Program of China under Grant No. 2018YFB1004300, the National Natural Science Foundation of China under Grant No. U1836206, U1811461, 61773361, the Project of Youth Innovation Promotion Association CAS under Grant No. 2017146.
This work is also partly supported by the funding of WeChat cooperation project.
\end{acks}

\bibliographystyle{ACM-Reference-Format}
\bibliography{main}
\end{document}